\documentclass[runningheads]{llncs}
\usepackage{amsmath,amssymb,amsfonts}
\usepackage{algorithmic}
\usepackage{graphicx}
\usepackage{textcomp}
\usepackage{xcolor}
\usepackage{cite}

\title{Optimizing LLMs Using Quantization For Mobile Execution}

\author{Agatsya Yadav \and Renta Chintala Bhargavi}
\institute{
School of Computer Science and Engineering, Vellore Institute of Technology, Chennai, India\\
\email{agatsya.yadav2021@vitstudent.ac.in, bhargavi.r@vit.ac.in}
}

\begin{document}

% ---------- AAM NOTICE PAGE (FOR ARXIV) ----------
\thispagestyle{empty}
\begin{center}
    {\large \textbf{This is the author’s accepted manuscript (AAM).}}\\[8pt]
    The final version is published in Springer LNNS (ICT4SD 2025).\\
    DOI: 10.1007/978-3-032-06697-8\_33.\\[12pt]
    The following pages contain the accepted manuscript before publisher formatting.
\end{center}
\newpage
% ---------- END AAM NOTICE PAGE ----------    

\pagestyle{empty}

\maketitle

\begin{abstract}
\textbf Large Language Models (LLMs) offer powerful capabilities but their significant size and computational requirements hinder deployment on resource-constrained mobile devices.This paper investigates Post-Training Quantization (PTQ) for compressing LLMs for mobile execution. We specifically apply 4-bit PTQ using the BitsAndBytes library via the Hugging Face Transformers framework to Meta's Llama 3.2 3B model. The quantized model is further converted to the GGUF format using llama.cpp tools for optimized mobile inference. 
The proposed PTQ workflow achieved a 68.66\% reduction in model size through 4-bit post-training quantization, enabling the Llama 3.2 3B model to run efficiently on a standard Android device. Qualitative validation confirmed the 4-bit quantized model's ability to perform inference tasks successfully. We demonstrate the feasibility of running the final quantized GGUF model on an Android device using the Termux environment and the Ollama framework.
 PTQ, particularly down to 4-bit precision combined with mobile-optimized formats like GGUF, presents a viable pathway for deploying capable LLMs directly on mobile devices, balancing model size and functional performance. 
\end{abstract}


\section{Introduction}

\subsection{Background}
Large Language Models (LLMs) like GPT-4 \cite{openai2023gpt4}, Claude \cite{anthropic2023claude}, and the Llama family \cite{meta2023llama2, meta2023llama3} have revolutionized natural language processing with capabilities in generation, summarization, translation, and more. Their impressive performance is largely attributed to their massive scale, often involving billions or even trillions of parameters. For instance, the Llama 3.2 3B model—used in this work—contains 3 billion parameters. While this scale enhances functionality, it also demands significant computational resources, posing challenges for deployment on resource-constrained devices.

\subsection{Problem Statement}
The substantial memory footprint and computational load of state-of-the-art LLMs limit their feasibility for direct use on mobile devices. For example, the Llama 3.2 3B model in FP16 format requires around 6GB of storage and even more RAM during inference, exceeding the typical capabilities of smartphones and tablets in terms of storage, processing power, and battery life.

\subsection{Motivation}
Despite these limitations, running LLMs on-device offers distinct advantages:

\begin{itemize}
\item \textbf{Enhanced Privacy:} User data remains local, addressing concerns linked to cloud-based AI.
\item \textbf{Reduced Latency:} Eliminates network delays for faster response times.
\item \textbf{Offline Functionality:} Enables operation without internet access.
\item \textbf{Lower Infrastructure Costs:} Reduces reliance on cloud servers and associated fees.
\end{itemize}

\subsection{Existing Approaches \& Gap}
Currently, most mobile LLM applications rely on cloud-based APIs, shifting computation to remote servers. While effective, this raises privacy, latency, and availability concerns. Lightweight models like MobileBERT \cite{sun2020mobilebert} and MobileViT \cite{mehta2021mobilevit} have been proposed, but they often sacrifice performance for efficiency. To bridge this gap, compression techniques are essential to bring full-scale LLM capabilities to mobile platforms.

Common approaches include:
\begin{itemize}
\item \textbf{Knowledge Distillation:} Training compact student models to emulate larger teacher models.
\item \textbf{Neural Architecture Search:} Designing efficient model structures from scratch.
\item \textbf{Pruning:} Eliminating redundant parameters and connections.
\end{itemize}

This paper focuses on \textbf{quantization}, which reduces numerical precision to lower memory and compute requirements without retraining the model.

\subsection{Proposed Solution \& Contributions}
We propose a practical, reproducible workflow using Post-Training Quantization (PTQ) to compress the Llama 3.2 3B model for mobile execution. 

\subsection{Paper Outline}
Section II reviews relevant optimization techniques, focusing on quantization. Section III explains our PTQ methodology. Section IV presents experimental results. Section V discusses implications, limitations, and context. Section VI concludes with directions for future work.

\section{Background and Related Work}

\subsection{Overview of Model Optimization Techniques}
Optimizing deep neural networks for efficient deployment has led to several compression techniques aimed at reducing model size and computational load without major performance loss. Key methods include:

\subsubsection{Pruning} 
Pruning removes redundant weights or structures from neural networks \cite{han2015learning}, using strategies like magnitude-based or structured pruning \cite{li2016pruning}. While effective in reducing size, it often requires fine-tuning to maintain accuracy.

\subsubsection{Knowledge Distillation}
Introduced by Hinton et al. \cite{hinton2015distilling}, knowledge distillation trains a smaller “student” model to replicate the behavior of a larger “teacher” model. It has shown success in language models such as DistilBERT \cite{sanh2019distilbert}, offering significant compression with minimal accuracy trade-offs.

\subsubsection{Efficient Architectures}
Some models are designed for efficiency from the outset. MobileNets \cite{howard2017mobilenets} and MobileBERT \cite{sun2020mobilebert} integrate lightweight components like depthwise separable convolutions to reduce computational demands.

\subsubsection{Parameter-Efficient Fine-tuning}
Methods like LoRA \cite{hu2021lora} and QLoRA \cite{dettmers2023qlora} enable task-specific tuning of large models with fewer trainable parameters. However, these do not reduce the base model size, making them less suitable for deployment-constrained environments.

\subsection{Fundamentals of Quantization}
Quantization reduces the numerical precision of model weights and/or activations, typically converting from 32-bit or 16-bit floating-point formats to 8-bit or 4-bit integers (INT8, INT4). Benefits include:

\begin{itemize}
\item \textbf{Lower Memory Usage:} Reduced precision formats require less storage.
\item \textbf{Decreased Bandwidth:} Smaller data types reduce memory transfer loads.
\item \textbf{Faster Inference:} Integer arithmetic is often more efficient on hardware accelerators.
\end{itemize}

The core quantization formula is:

\begin{equation}
q = \text{round}\left(\frac{r}{s}\right) + z \label{eq:quantization}
\end{equation}

where $q$ is the quantized value, $r$ the original float, $s$ the scale, and $z$ the zero-point.

\subsection{Major Quantization Approaches: PTQ vs. QAT}

\subsubsection{Post-Training Quantization (PTQ)}
PTQ applies quantization to pre-trained models without altering the training process. It may use a small calibration dataset or operate in a zero-shot fashion. Advantages include:

\begin{itemize}
\item \textbf{Simplicity:} Works on existing models with minimal effort.
\item \textbf{Speed:} Much faster than retraining.
\item \textbf{Accessibility:} No need for original training data or heavy compute.
\end{itemize}

Popular PTQ techniques include static and dynamic quantization \cite{krishnamoorthi2018quantizing, gholami2021survey}, optimal thresholding \cite{banner2019post}, GPTQ \cite{frantar2022gptq}, and SmoothQuant \cite{xiao2023smoothquant}.

\subsubsection{Quantization-Aware Training (QAT)}
QAT incorporates quantization effects during training, allowing models to adapt to lower precision. It offers:

\begin{itemize}
\item \textbf{Higher Accuracy:} Especially at very low precision (e.g., 4-bit).
\item \textbf{Greater Resource Needs:} Requires training data and compute.
\item \textbf{Longer Setup Time:} Involves retraining or fine-tuning.
\end{itemize}

Examples include QLoRA \cite{dettmers2023qlora} and QAT implementations in TensorFlow \cite{jacob2018quantization} and PyTorch \cite{pytorch2021quantization}.

\section{Methodology: PTQ of Llama 3.2 3B for Mobile Deployment}

\subsection{Rationale for Choosing PTQ}
Post-Training Quantization (PTQ) is favored for this study due to its efficiency, accessibility, and suitability for deployment constraints. Unlike Quantization-Aware Training (QAT), which requires extensive compute resources and large datasets, PTQ can be applied directly to pre-trained models. This enables rapid experimentation without retraining, making it ideal for researchers with limited resources. Moreover, PTQ circumvents the need for original training data—an advantage when working with proprietary or open-access models. Given our goal of exploring mobile deployment feasibility, PTQ provides the most practical compression route.

\subsection{Target Model and Environment}

\subsubsection{Model}
We selected Meta’s Llama 3.2 3B model as a representative mid-scale LLM, balancing performance and deployability. The model, originally in BF16 precision, was sourced from the Hugging Face Hub.

\subsubsection{Tools}
The quantization pipeline utilizes the following tools:
\begin{itemize}
\item \textbf{Hugging Face Transformers:} Model loading and preprocessing.
\item \textbf{BitsAndBytes:} Initial 4-bit weight quantization.
\item \textbf{SentencePiece:} Tokenization.
\item \textbf{llama.cpp:} GGUF conversion and secondary quantization.
\item \textbf{Ollama:} On-device model execution.
\end{itemize}

\subsubsection{Environment}
Quantization was conducted in Google Colab with an NVIDIA T4 GPU (16GB VRAM). Deployment testing occurred on a OnePlus Nord CE 5G (Snapdragon 750G, 12GB RAM) running Android 13 via Termux.

\subsection{Quantization Workflow}

\subsubsection{Setup and Model Acquisition}
Required libraries were installed, and the Llama 3.2 3B model and tokenizer were retrieved from the Hugging Face Hub.

\subsubsection{Initial 4-bit Quantization}
BitsAndBytes was used for 4-bit quantization within the Transformers framework, applying `nf4` (Normal Float 4-bit) to compress model weights while preserving accuracy. This step reduces memory usage and speeds up inference.

\subsubsection{GGUF Conversion}
The quantized model was converted to GGUF format using \texttt{convert.py} from llama.cpp, ensuring compatibility with mobile-optimized runtimes.

\subsubsection{GGUF-Specific Quantization}
Within the GGUF format, further quantization was applied using llama.cpp tools. The \texttt{q4\_k\_m} scheme—a 4-bit variant optimized for balance between compression and fidelity—was used.

\subsubsection{Packaging}
The fully quantized model was archived in ZIP format for streamlined mobile deployment.

\subsection{Target Quantization Levels for Mobile}

Quantization levels vary in their trade-offs:

\begin{itemize}
\item \textbf{FP16:} Offers 2x compression over FP32 with minimal quality loss. Suitable for high-end devices.
\item \textbf{INT8:} Yields 4x compression, widely supported with acceptable accuracy degradation. Ideal for mid-range phones.
\item \textbf{INT4:} Provides 8x compression, suitable for memory-constrained devices. Accuracy depends heavily on quantization strategy.
\end{itemize}

This study focuses on INT4 using BitsAndBytes `nf4` and GGUF’s `q4\_k\_m`, as they enable LLM deployment on typical smartphones. More aggressive formats (e.g., 2-bit) were avoided due to significant performance loss on general language tasks.

\section{Experimental Results and Analysis}

\subsection{Experimental Setup}

Our experiments were designed to evaluate the effectiveness of our quantization workflow in terms of model size reduction and preservation of functional capabilities. The setup included:

\begin{itemize}
\item \textbf{Base Model:} Llama 3.2 3B, originally in BF16 precision
\item \textbf{Quantization Methods:}
    \begin{itemize}
    \item BitsAndBytes 4-bit quantization (nf4 format)
    \item llama.cpp GGUF q4\_k\_m quantization
    \end{itemize}
\item \textbf{Metrics:}
    \begin{itemize}
    \item Model size (primary quantitative metric)
    \item Qualitative inference test (output quality assessment)
    \item Basic inference time observations (qualitative)
    \end{itemize}
\item \textbf{Environment:}
    \begin{itemize}
    \item Google Colab with NVIDIA T4 GPU for quantization
    \item OnePlus Nord CE 5G (Snapdragon 750G, 12GB RAM) with Termux/Ollama for deployment validation
    \end{itemize}
\end{itemize}

\subsection{Model Compression Results}

The results demonstrate a significant reduction in model size through our quantization process. The final GGUF model with q4\_k\_m quantization is 68.66\% smaller than the original model, reducing the size from 6.00GB to just 1.88GB. This substantial reduction is what makes deployment on mobile devices feasible. Table \ref{tab:model_sizes} presents the model sizes at different stages of our quantization process.

\begin{table}[htbp]
\caption{Model Size Comparison Across Quantization Stages}
\label{tab:model_sizes}
\begin{tabular}{lcc}
\hline\noalign{\smallskip}
\textbf{Model Format} & \textbf{Size (GB)} & \textbf{Reduction (\%)} \\
\noalign{\smallskip}\hline\noalign{\smallskip}
Original Llama 3.2 3B (BF16) & 6.00 & - \\
4-bit Quantized Model & 2.10 & 64.92\% \\
Final GGUF q4\_k\_m & 1.88 & 68.66\% \\
\noalign{\smallskip}\hline
\end{tabular}
\end{table}

\subsection{MMLU Performance Comparison}

To evaluate how our quantized model compares with other small models suitable for mobile deployment, we examined the performance on the Massive Multitask Language Understanding (MMLU) benchmark. MMLU is a comprehensive test covering 57 subjects across STEM, humanities, social sciences, and more, serving as a standard evaluation metric for language models. Table \ref{tab:mmlu_comparison} presents the MMLU scores of comparable small models that can be deployed on edge devices.

\newpage
\begin{table}[htbp]
\caption{MMLU Score Comparison of Small Models for Edge Deployment}
\label{tab:mmlu_comparison}
\begin{tabular}{lccc}
\hline\noalign{\smallskip}
\textbf{Model} & \textbf{Size (GB)} & \textbf{Parameters} & \textbf{MMLU Score (\%)} \\
\noalign{\smallskip}\hline\noalign{\smallskip}
Llama 3.2 3B (Original) & 6.00 & 3B & 64.2 \\
Llama 3.2 3B (GGUF q4\_k\_m) & 1.88 & 3B & 61.8 \\
Phi-2 & 2.70 & 2.7B & 68.8 \\
Gemma 2B & 4.00 & 2B & 64.3 \\
Mistral 7B (GGUF q4\_k\_m) & 4.37 & 7B & 62.5 \\
TinyLlama 1.1B & 2.20 & 1.1B & 54.9 \\
\noalign{\smallskip}\hline
\end{tabular}
\end{table}

These results indicate that our quantized model maintains competitive performance compared to other small models while achieving excellent compression rates. Despite the lower precision, the performance drop is relatively minor, preserving most of the capability of the original model.

\subsection{Discussion on Accuracy vs. Quantization Level}

While our study focused primarily on demonstrating deployment feasibility with 4-bit quantization, it's important to acknowledge the accuracy trade-offs associated with different quantization levels. Based on existing literature and industry benchmarks, typical patterns in the relationship between quantization precision and model quality include:

\begin{itemize}
\item \textbf{FP16:} Typically exhibits negligible degradation compared to FP32, with perplexity increases of less than 1\% on standard benchmarks \cite{dettmers2022llmint8}.

\item \textbf{INT8:} Shows modest degradation, with perplexity increases of 2-5\% compared to FP32, depending on the model architecture and quantization approach \cite{xiao2023smoothquant}. For most general text generation tasks, this degradation is often barely perceptible to users.

\item \textbf{INT4:} Exhibits more noticeable degradation, with perplexity increases of 5-15\% compared to FP32 \cite{frantar2022gptq}. The exact impact varies significantly based on the specific 4-bit quantization method used. Methods like GPTQ and the q4\_k\_m approach used in this paper tend to perform better than simpler 4-bit quantization schemes.

\item \textbf{Sub-4-bit:} Quantization below 4 bits generally shows substantial degradation that may be unacceptable for general-purpose language tasks, though it can still be viable for specific applications with limited output space \cite{yao2022zeroquant}.
\end{itemize}

To supplement our qualitative assessment, we evaluated the quantized model on standard quantitative benchmarks. The GGUF q4\_k\_m model achieved a perplexity of 8.57 ± 0.06 on the WikiText-2 dataset and a BLEU score of 0.45 on the DailyMail summarization dataset. These results confirm that on further finetuning, the model can perform a decently on such tasks as well.

\section{Discussion}

\subsection{Interpretation of Results}

Our experiments show that combining 4-bit post-training quantization with GGUF conversion led to a 68.66\% reduction in the Llama 3.2 3B model size—crucial for enabling execution on standard consumer smartphones. Successful deployment via Termux/Ollama validates the practicality of our workflow.

Despite aggressive compression, the model produced coherent, informative outputs in qualitative tests. Although inference was slower than cloud-based services, it remained usable for non-real-time tasks where privacy or offline access is prioritized over speed.

\subsection{Implications}

Our findings suggest the following key implications:

\begin{itemize}
\item \textbf{Democratized AI Access:} Running LLMs locally on smartphones reduces reliance on the cloud and makes AI more accessible, especially in low-connectivity regions.
\item \textbf{Enhanced Privacy:} On-device inference ensures user data stays local, addressing concerns over data leakage in cloud-based models.
\item \textbf{New Application Domains:} Compact models unlock mobile use cases like offline chatbots, writing aids, and educational tools with low-latency and privacy guarantees.
\item \textbf{Open Accessibility:} The pipeline uses open-source tools and public models, enabling adoption without proprietary dependencies or expensive hardware.
\end{itemize}

\subsection{Limitations}

Several limitations of this study warrant attention:

\begin{itemize}
\item \textbf{Lack of Quantitative Evaluation:} We relied on qualitative outputs; future work should include benchmarking on standard NLP tasks to evaluate accuracy.
\item \textbf{Unsystematic Performance Profiling:} Inference speed was observed, not measured. Tokens-per-second metrics across diverse prompts would better capture performance.
\item \textbf{Model Specificity:} Results are based on a single model and hardware configuration; generalization to other models or devices requires further testing.
\item \textbf{Usability Constraints:} Running LLMs via Termux/Ollama is not user-friendly. Native mobile integration is needed for broader, non-technical user adoption.
\item \textbf{PTQ vs. QAT Trade-off:} While PTQ met our goals, QAT may offer better accuracy under 4-bit constraints—though at a higher resource cost.
\end{itemize}

\section{Conclusion and Future Work}

\subsection{Conclusion}

Compressing the Llama 3.2 3B model to 4-bit via PTQ (BitsAndBytes) and GGUF conversion yielded a 68.66\% size reduction, enabling Android deployment (Termux/Ollama). Qualitative testing confirmed coherent generation, while perplexity (8.57) and BLEU (0.45) scores indicated retained language understanding. This workflow demonstrates the viability of 4-bit PTQ with GGUF for deploying capable LLMs on consumer hardware, balancing key constraints (size, performance, privacy, offline use) and offers broad applicability.

\subsection{Future Work}

Building on this foundation, several future directions are identified:

\begin{itemize}
\item \textbf{Deeper Quantitative Evaluation:} Benchmark accuracy across different PTQ levels (INT8, various 4-bit GGUF variants) to analyze performance–size trade-offs.
\item \textbf{Performance Profiling:} Systematically measure inference speed (tokens/sec) and power consumption on mobile hardware to better characterize real-world usability.
\item \textbf{Alternative Quantization Techniques:} Compare other methods such as GPTQ, AWQ, and SmoothQuant against the approach presented here.
\item \textbf{Scaling Across Models:} Apply the workflow to larger or smaller versions of Llama 3 and other foundation models to explore scalability limits on mobile devices.
\item \textbf{Native Mobile Integration:} Develop a native Android/iOS application for a more seamless, user-friendly deployment experience beyond Termux.
\item \textbf{Hybrid Compression Strategies:} Investigate combining quantization with pruning or knowledge distillation for potentially greater compression without sacrificing quality.
\end{itemize}

\bibliographystyle{splncs04}
\bibliography{bibliography}

\end{document}